# Data Engineering for the Analysis of Semiconductor Manufacturing Data

Peter Turney
Knowledge Systems Laboratory
Institute for Information Technology
National Research Council Canada
Ottawa, Ontario, Canada
K1A 0R6

613-993-8564 (voice)
613-952-7151 (fax)
peter@ai.iit.nrc.ca

**Abstract**

We have analyzed manufacturing data from several different semiconductor manufacturing plants, using decision tree induction software called Q-YIELD. The software generates rules for predicting when a given product should be rejected. The rules are intended to help the process engineers improve the yield of the product, by helping them to discover the causes of rejection. Experience with Q-YIELD has taught us the importance of *data engineering* — preprocessing the data to enable or facilitate decision tree induction. This paper discusses some of the data engineering problems we have encountered with semiconductor manufacturing data. The paper deals with two broad classes of problems: engineering the features in a feature vector representation and engineering the definition of the target concept (the classes). Manufacturing process data present special problems for feature engineering, since the data have multiple levels of granularity (detail, resolution). Engineering the target concept is important, due to our focus on understanding the past, as opposed to the more common focus in machine learning on predicting the future.

## 1. Introduction

We define *data engineering* as the transformation of raw data into a form useful as input to algorithms for inductive learning.[1] This paper is concerned with the transformation of semiconductor manufacturing data for input to a decision tree induction algorithm. We have been using decision tree induction for process optimization. The optimization task that we address is to discover what aspects of a manufacturing process are responsible for a given class of rejected products. A product in semiconductor manufacturing may be considered to be an integrated circuit, a wafer (a disk of silicon, usually holding about 100 to 1,000 integrated circuits), or a batch of wafers (usually about 20 to 30 wafers). A product is usually accepted or rejected on the basis of electrical measurements. (See Van Zant (1986) for a good introduction to semiconductor manufacturing.)

We have analyzed data from several different semiconductor manufacturing plants. The data were analyzed using Q-YIELD, which generates rules for predicting when a given product will be rejected, given certain process measurements (Famili & Turney, 1991; 1992).[2] The rules are intended to help process engineers improve the yield of the product, by helping them discover the causes of rejection. In general, there are two types of applications for inductive learning algorithms: they may be used to predict the future or to understand the past. Our emphasis has been on understanding the past. This places certain constraints on the software that we use. For example, it is very important that the induced model should be readily understandable by process engineers, which excludes neural network models.

This real-world application of machine learning has presented us with some interesting technical problems, which do not seem to have been considered in the machine learning literature. Section 2 discusses the problems of engineering the features. Manufacturing process data have multiple levels of granularity (levels of detail; levels of resolution), which makes the data difficult to represent in the standard feature vector notation. Section 3 discusses the problems of engineering the classes. Most papers on inductive learning assume that the definition of the target concept is given. In our work, determining the best definition of the target concept is a large part of the task. For each problem we describe, we outline the solutions we have adopted and the open questions. In general, we have not had the resources to validate our solutions by comparing them with alternative approaches.

The conclusion is that data engineering is essential to the successful application of decision tree induction to semiconductor manufacturing. Data engineering is currently much more an art than a science. We present here a list of recipes for data engineering with semiconductor manufacturing data, but what we need is a unifying scientific foundation for this diverse collection of recipes.

Data engineering presents interesting philosophical issues: How can we make sure we do not massage the data too much and see only what we want to see? Is it better to invent new techniques for data engineering for existing learning algorithms or to invent new learning algorithms that do not require data engineering? Is data engineering inherently limited to the non-scientific collection of *ad hoc* recipes and techniques? We do not have answers for these questions.[3]

## 2. Data Engineering for Defining the Features

In this section, we examine the problems we have encountered with engineering of features. Manufacturing process data do not fit painlessly into the feature vector representation that is used by most inductive learning algorithms.

### 2.1 Multiple Levels of Granularity

There are (at least) four levels of granularity in semiconductor process data:

1. **IC-Level:** At the lowest (finest, most detailed) level of granularity, there are individual integrated circuits (ICs).[4]

2. **Site-Level:** The next level of granularity is the test site. ICs are manufactured on disks of silicon, about 12 centimeters in diameter. These disks are called wafers. Each wafer typically has five test sites. A test site is a location on the wafer where a simple circuit, with well-known properties, is constructed, instead of the more complex circuits that fill the remainder of the wafer. These simple circuits are used to test the manufacturing process. Since the test sites are well-understood, they facilitate diagnosis of the manufacturing process.

3. **Wafer-Level:** At the next level, there is a wafer. A wafer is a disk of silicon, usually holding about 100 to 1,000 ICs.

4. **Batch-Level:** At the top level, there is a batch of wafers. A batch usually consists of about 20 to 30 wafers. The wafers in a batch stay close together for most of the manufacturing process, so they often have many properties in common (such as similar types of defects).[5]

There may be 30,000 ICs in a batch (e.g., 30 wafers with 1,000 ICs each). Thus each measurement made at the batch-level of granularity may correspond to 30,000 measurements made at the IC-level of granularity. This shows that these four levels of granularity span a significant range. In some of our analyses, we only deal with two or three of these four levels.

Typically from 10 to 100 measurements are made at each level of granularity. Batch-level and wafer-level measurements deal with aspects of the manufacturing process. For example, when a batch of wafers is baked in an oven, one of the batch-level measurements might be the maximum oven temperature. There is no reason to record this information at the site-level (for example), since all the site-level temperature measurements would be (approximately) identical within a given batch. Site-level and IC-level measurements are typically electrical measurements that are used for quality control. A wafer may be rejected if certain site-level measurements are not within bounds. An IC may be rejected if certain IC-level measurements are not within bounds. Part of the reason that granularity is important is that decisions (accept or reject) are made at different levels of granularity (reject a whole wafer or reject a single IC).

The data at these different levels of granularity are frequently stored in separate databases. Data from the manufacturing process are recorded at the batch-level in one database, while data from electrical testing are recorded at the IC-level in a second database. This introduces the mundane difficulty of extracting data from two or more databases, but there is also a challenging data engineering problem: The data have a structure that is not naturally represented in the feature vector format required by most decision tree induction algorithms.

So far, our practice has been to convert the data to a feature vector format by moving all measurements up to the highest level of granularity that is relevant for the given manufacturing problem (often the batch-level). We have experimented with two methods for transforming lower-level data to higher-level data:

**Method A:** Suppose that we wish to move site-level data up to the batch-level. Let $X$ be an electrical parameter (the voltage drop across a diode, for example) that is measured at five test sites on a wafer. In a batch of 24 wafers, there will be 120 ($24 \times 5 = 120$) measurements of $X$. To bring $X$ up from the site-level to the batch-level, we can introduce five new batch-level features:

1. the average of $X$ in the 120 measurements in the given batch
2. the standard deviation of $X$ in the 120 measurements in the given batch
3. the median of $X$ in the 120 measurements in the given batch
4. the minimum of $X$ in the 120 measurements in the given batch
5. the maximum of $X$ in the 120 measurements in the given batch

This can result in a large feature vector, since every lower-level measurement results in five higher-level features. It can also result in a shortage of cases. A database with 1,200 records (cases, examples), where each field (attribute, feature, measurement) is measured at the site-level, yields 10 batch-level feature vectors ($1,200/120 = 10$). Thus an abundance of data is transformed into a shortage of data. However, if the manufacturing problem is due to fluctuations in the process at the batch-level, then the apparent abundance of data was an illusion, since the database only has a small amount of information about batch-level fluctuations.

**Method B:** The electrical measurements at the site-level are generally related to the decision to accept or reject the wafer. Suppose that a wafer is rejected when two or more of the five measurements of electrical parameter $X$ are above a threshold $T$. To bring $X$ up from the site-level to the batch-level, we can introduce a new variable $Y$ defined as the percentage of the 24 wafers for which two or more of the five measurements of electrical parameter $X$ are above the threshold $T$. This new variable $Y$ now has the same level of granularity as the batch-level process data.

We prefer Method B to Method A, since the new variable $Y$ is easier for the process engineers to understand than the five variables in Method A. However, Method B does not always work, since not all site-level measurements are directly tied to the decision to accept or reject the wafer: some measurements may be recorded purely for the information they provide about the manufacturing process.

These two methods for changing granularity usually work for us, but a number of open questions arise:

1. What happens if we bring the batch-level data down to the site-level, instead of bringing the site-level data up to the batch-level?
2. Is there an algorithm that can handle the raw data, without preprocessing the data to either reduce or increase the level of granularity? This appears to be an ideal problem for Inductive Logic Programming (Lavrac & Dzeroski, 1994).
3. Are there better ways of changing granularity?

The issue of granularity is an interesting research problem. It will be a common problem in any discrete industrial manufacturing process (a discrete process involves individual components, in contrast to a continuous process, such as petroleum refining). It is somewhat surprising that it has not received more attention.

## 2.2 Deciding What Information to Record or Analyze

With semiconductor manufacturing data, we are frequently faced with several hundred potential features in our feature vectors. Our rule of thumb is to use every feature that is possibly relevant for the given target class. We leave it to the decision tree induction algorithm to select the best subset from the set of features that are given as input. This is perhaps overloading the decision tree induction algorithm, but it appears to work (most of the time).

Since we rely on the decision tree induction algorithm to ignore irrelevant features, our task is to attempt to include all possibly relevant features. We list here some potential features that might be overlooked, for the benefit of those who intend to apply inductive learning to manufacturing data:

1. It can be useful to record the name or employee ID of the operator of the machinery, when there is manually operated machinery in the manufacturing

    process. Some of the problems in the process can be due to inexperienced operators or variation in the operators' methods.

2. When several different machines perform the same task in parallel, it can be useful to record for each product which machine was used. One of the machines might have a defect, which can be exposed by a rule of the form, "If machine number 3 is used, the wafer will be rejected."

3. It might be useful to record the name of the supplier for raw materials, when there are several suppliers. For example, a rule of the form, "If the silicon wafer comes from supplier Z, then the wafer will be rejected," might suggest that the wafers from supplier Z have impurities.

4. It might be useful to record environmental conditions, such as humidity.

5. It is often useful to record time. This is discussed in the next section.

Part of data engineering is deciding what data to supply to the induction algorithm. To some extent, this is automated with decision tree induction, since we can put everything in and let the algorithm select the relevant subset. However, this approach has the disadvantage that the speed of the algorithm decreases as the number of features increases.

### 2.3 Representing Time

Many aspects of a manufacturing process vary with time. Time can sometimes be a surrogate for important variables that are not recorded. Suppose the yield in a certain process is determined by the skill of the operators, but the operator ID is not recorded. However, perhaps the less skilled workers are assigned to the night shift, so the yield will be correlated with the time of day.

    There are many ways to represent time in a feature vector format. If it is possible that the target class has a cyclical pattern, then time should be represented in a way that encourages the discovery of cyclical patterns. The features could be the hour of the day, the day of the week, and the week of the month. An extra feature might be added to flag weekends or holidays, if the manufacturing process operates differently on these days.

    It is possible that a unique event suddenly caused the manufacturing process to go awry. In this case, we want to emphasize the sequential nature of time, rather than its cyclical nature. Often a database record has a time-stamp that records the year, month, day, hour, and minute of a measurement. When we want to emphasize the sequential nature of time, we convert these five numbers into a single feature, such as "minutes from midnight, January 1, 1990". If a rule makes use of this feature, we convert it back into a more comprehensible format for reporting purposes.

    When time is not explicitly recorded in a database, it is sometimes possible to

extract sequential order from the batch ID. Most plants assign an ID to each batch and the IDs are often a combination of digits that are assigned sequentially. This ordering information is sufficient to detect unique events — it may be unnecessary to know the absolute time of an event.

### 2.4 Other Feature Engineering Problems

There are some common feature engineering problems that we have not discussed here, because they are not specific to manufacturing data. We briefly mention three of these problems. (1) Occasionally there are missing feature values in the data. We simply throw out all cases that have any missing values. (2) Bad sensor readings can cause outliers. To handle this problem, we have an upper and lower limit for each sensor reading. Cases are flagged and discarded when any feature value is outside of its limits. (3) Often the features are highly correlated with each other. This confuses the analysis, since highly correlated features can act as "synonyms" in the decision tree. When the decision tree induction algorithm is run on several batches of data from the same process, minor variations in the data can cause radically different trees to be generated. We screen the data for highly correlated features by generating a table of correlations for all pairs of features.

## 3. Data Engineering for Defining the Target Class

The *yield* of a process is a continuous variable but decision tree induction usually requires a discrete variable for the class. Some of the problems we have encountered come from attempting to convert a continuous variable into a discrete variable. This section examines some of these problems. We should begin by explaining why we convert a continuous variable into a discrete variable.

In statistics, induction with discrete classes is called *classification* or *discrimination*. Induction with a continuous dependent variable is called *regression* or *curve-fitting*. Although decision tree induction is usually applied to classification, it can be applied to regression, as demonstrated by CART (Breiman *et al.*, 1984). Our goal is to produce easily understood rules, to assist the process engineers. We believe that regression trees are harder to understand than classification trees, so we prefer to convert the continuous dependent variable into a discrete class, instead of applying regression trees.

### 3.1 Defining the Target Class

The *target class* is the class that is to be learned by the induction algorithm. In our work, the target class is the manufacturing problem that interests the process engineer. Usually the process engineer is concerned with low yield. The *yield* of a process is the percentage of ICs or wafers that are acceptable. Suppose that the yield of a process is usually above 90% but sometimes dips below 90% and the process engineer wants to understand what is causing the dip. In the simplest case, there is a batch-level measurement called *yield* and the target class is "*yield* is less than 90%". We can define the target class as a symbolic variable with the value 1 when the yield is below 90% and the value 0 when the yield is above 90%.

We convert the continuous *yield* variable to a discrete variable using a threshold, such as 90%. There are (at least) three ways to set a threshold for the yield. (1) We may use external factors (economic factors, management decisions, pressure from competition) to determine the desired minimum yield for the process; (2) we can choose the median yield, so that we have a balance of examples and counter-examples; or (3) we can look at the data to see whether there is a natural threshold, based on clusters in the data. We find that we tend to get better results with approaches (2) and (3), rather than (1). We often experiment with several different thresholds. We use visual aids to suggest possible thresholds. One aid is a histogram of the yield (the $x$ axis is the yield and the $y$ axis is the number of batches with the given yield). Sometimes there will be a valley in the histogram that suggests a natural place for a threshold. Another aid is a plot of the yield over time (the $x$ axis is time and the $y$ axis is yield). Sometimes there are recurrent dips in the plot that can readily be isolated with the right threshold.

The *yield* of a process is a composite variable, since there are many different reasons for rejection of a wafer or IC. In a process with a yield of 90%, there may be 30 different classes of problems in the 10% of parts that are rejected. Treating each problem separately can make the task simpler for the induction algorithm. Suppose that electrical measurements are made at five test sites on a wafer and a wafer is rejected when two or more of the five measurements of electrical parameter $X$ are above a threshold $T$. This electrical measurement $X$ is one way that a wafer can be rejected and we can focus on $X$ instead of examining *yield*. To bring $X$ up from the wafer-level to the batch-level, we can introduce a new variable $Y$ defined as the percentage of the wafers for which two or more of the five measurements of electrical parameter $X$ are above the threshold $T$ (as we discussed in Section 2.1). We can then define two classes of batches, those for which $Y$ is below some threshold $U$ and those for which $Y$ is above $U$. The target class is "$Y$ is below $U$". The same issues arise in setting the threshold $U$ as arose in setting the threshold on the yield.

Some open questions are:

1. Can we automate the selection of a threshold in the definition of a target class?
2. Should we use regression trees instead of classification trees (Breiman *et al.*, 1984)? Is there a way to make regression trees easier to understand?

We have not investigated these issues.

### 3.2 Grey Cases

As we described in Section 3.1, we typically define the target class by applying a threshold to a continuous variable. Some cases are clearly instances of the target class ("black") and some cases are clearly not instances of the target class ("white"), but there are frequently borderline cases ("grey") that can confuse the induction algorithm. We have found that we can occasionally improve the performance of the induction algorithm by defining a "grey region" around the threshold that defines the target class. We delete the cases in the grey region (in both the training and the testing data).

It may be a dubious practice to delete cases when making predictions, but our focus is on understanding the cause of the manufacturing problem; the focus is not on prediction. We have found that dropping the grey region appears to enhance understanding. However, we have not rigorously tested this hypothesis.

Another issue is deciding the boundaries of the grey region. There is a trade-off between a large region, which increases the contrast between the examples and counter-examples, and a small region, which increases the amount of data available for the induction algorithm. We do not yet have a principled approach to setting the boundaries. A histogram can be helpful.

### 3.3 Other Class Engineering Problems

Our target classes can overlap, since a rejected part (wafer or IC) may have several different problems simultaneously. We handle this in the standard way, by generating a separate decision tree for each target class. Each tree has only two classes, target and non-target.

## 4. Conclusions

The above examples show that a significant amount of data engineering is involved in the application of decision tree induction to semiconductor manufacturing data. There are many open questions raised by our data engineering methods and many assumptions that have not yet been investigated. We believe that it is possible and worthwhile to build a firm theoretical foundation for data engineering. We are hopeful that the recipes and open questions raised here can contribute to such a foundation.

## Notes

1. This definition suggests that data engineering is always done by hand. We do not mean to exclude the possibility of automatic data engineering, but we have not been able to invent a more satisfying definition of data engineering.

2. Q-YIELD is a commercial product, available from Quadrillion Corporation, 380 Pinhey Point Road, Dunrobin, Ontario, Canada, K0A 1T0. The software is based on a prototype that was developed at the NRC.

3. These issues were raised by Joel Martin, in conversation.

4. For some tasks, it is reasonable to consider a lower level of granularity, such as the components (flip flops, transistors, gates) within an IC. The four levels listed here are not meant to be exhaustive.

5. There are higher levels of granularity, such as a production run, but we do not usually analyze the data at this level of granularity.

## Acknowledgments

Thanks to Michael Weider and Joel Martin for their very helpful comments on earlier versions of this paper.